\documentclass{article}
\usepackage{spconf,amsmath,graphicx}
\usepackage{multirow}
\usepackage{url}
\usepackage{enumitem}
\usepackage{textcomp}
\usepackage[table]{xcolor}

\title{ALTERNATING WEAK TRIPHONE/BPE ALIGNMENT SUPERVISION FROM HYBRID MODEL IMPROVES END-TO-END ASR}
\name{Jintao Jiang$^1$, Yingbo Gao$^2$, Mohammad Zeineldeen$^2$, Zoltan Tuske$^1$}
\address{$^1$AppTek, McLean, Virginia, USA \\ $^2$AppTek GmbH, 52062 Aachen, Germany \\
\small \texttt{\{jjiang|ygao|mzeineldeen|ztuske\}@apptek.com}}

\begin{document}

%
\maketitle
\begin{abstract}
In this paper, alternating weak triphone/BPE alignment supervision is proposed to improve end-to-end model training. Towards this end, triphone and BPE alignments are extracted using a pre-existing hybrid ASR system. Then, regularization effect is obtained by cross-entropy based intermediate auxiliary losses computed on such alignments at a mid-layer representation of the encoder for triphone alignments and at the encoder for BPE alignments. 
Weak supervision is achieved through strong label smoothing with parameter of 0.5.
Experimental results on TED-LIUM 2 indicate that either triphone or BPE alignment based weak supervision improves ASR performance over standard CTC auxiliary loss.
Moreover, their combination lowers the word error rate further. We also investigate the alternation of the two auxiliary tasks during model training, and additional performance gain is observed. Overall, the proposed techniques result in over 10\% relative error rate reduction over a CTC-regularized baseline system.
\end{abstract}
\vspace{-.5em}
\section{Introduction}
\vspace{-.5em}
\label{sec:intro}
End-to-end (E2E) automatic speech recognition (ASR) modeling has achieved incredible milestones in recent years \cite{prabhavalkar2023end,li2022recent,bahdanau2016end,lu2016training}.
The E2E modeling has been democratizing machine learning in ASR as researchers do not need to be 
exceptionally proficient at speaking, or writing a language.
At the same time, E2E has also been making multilingual ASR feasible and shortening the model development time as demonstrated by Whisper \cite{radford2022robust}.
For these reasons and beyond, researchers have been moving away from traditional hybrid ASR modeling \cite{Bourlard1993}.
Nevertheless, E2E approaches tend to suffer from spiky and/or inaccurate alignment problems \cite{Gakuto2019,mahadeokar2021}.
Thus, the more accurate hybrid alignments still represent valuable resources these days.

In our previous study \cite{jiang2023weak}, we demonstrated that an auxiliary loss using triphone alignment extracted by a hybrid system can indeed aid Attention-based Encoder Decoder (AED) E2E modeling.
Especially, if it is combined with weak supervision achieved through label smoothing \cite{szegedy2016rethinking}, and when the auxiliary loss is placed around the middle, to an intermediate layer representation, of the encoder.
Standard AED training usually includes joint CTC-attention training using the encoder output and the target sequence of the main task \cite{kim2017joint}.
Our previous study also confirmed that such auxiliary loss should be placed at the higher layers of the encoder, and that it is less efficient than the hybrid alignment based approach.


Inspired by these findings, we extend our previous study, and (1) combine multiple auxiliary losses using different target sequences (triphone and BPE) and different encoder layer outputs; (2) also replace the CTC loss with weak BPE alignment supervision at the encoder output; (3) we experiment with alternating the two losses during training. 



\vspace{-.5em}
\section{Related Work}
\vspace{-.5em}
\label{sec:relatedwork}
The major E2E architectures for speech recognition are Attention-based Encoder Decoder (AED) \cite{chan2016listen,zeyer2019comparison}, Connectionist Temporal Classification (CTC) \cite{graves2006connectionist,graves2014towards}, and Recurrent Neural Networks-Transducer (RNN-T) \cite{graves2012sequence,rao2017exploring,he2019streaming,jain2019rnn}; our study focuses on improving BPE based AED model \cite{sennrich2015neural}.
Multi-task learning has been widely exploited for E2E modeling \cite{kim2017joint,boyer2021study,luong2015multi,crawshaw2020multi,zhang2021survey,lee2021intermediate}.
In this paper, we aim at utilizing  pre-existing hybrid ASR models to aid the E2E modeling similar to \cite{liu2021improving,sogaard2016deep}, or \cite{toshniwal2017multitask} which used state alignment supervision in a multi-task training approach with an AED model.

The initial exploration of using label smoothing and weak triphone alignment supervision was proposed in our previous study \cite{jiang2023weak}.
Similar to our previous study, we use the recipe from \cite{zeyer2019comparison} which is built on RETURNN, a TensorFlow-based deep learning platform \cite{zeyer2018returnn}.
Nevertheless, we improve upon the original recipe to produce a new and stronger baseline.

\vspace{-.5em}
\section{Methodology}
\begin{figure}[h!]
    \centering
    \includegraphics[scale=0.48]{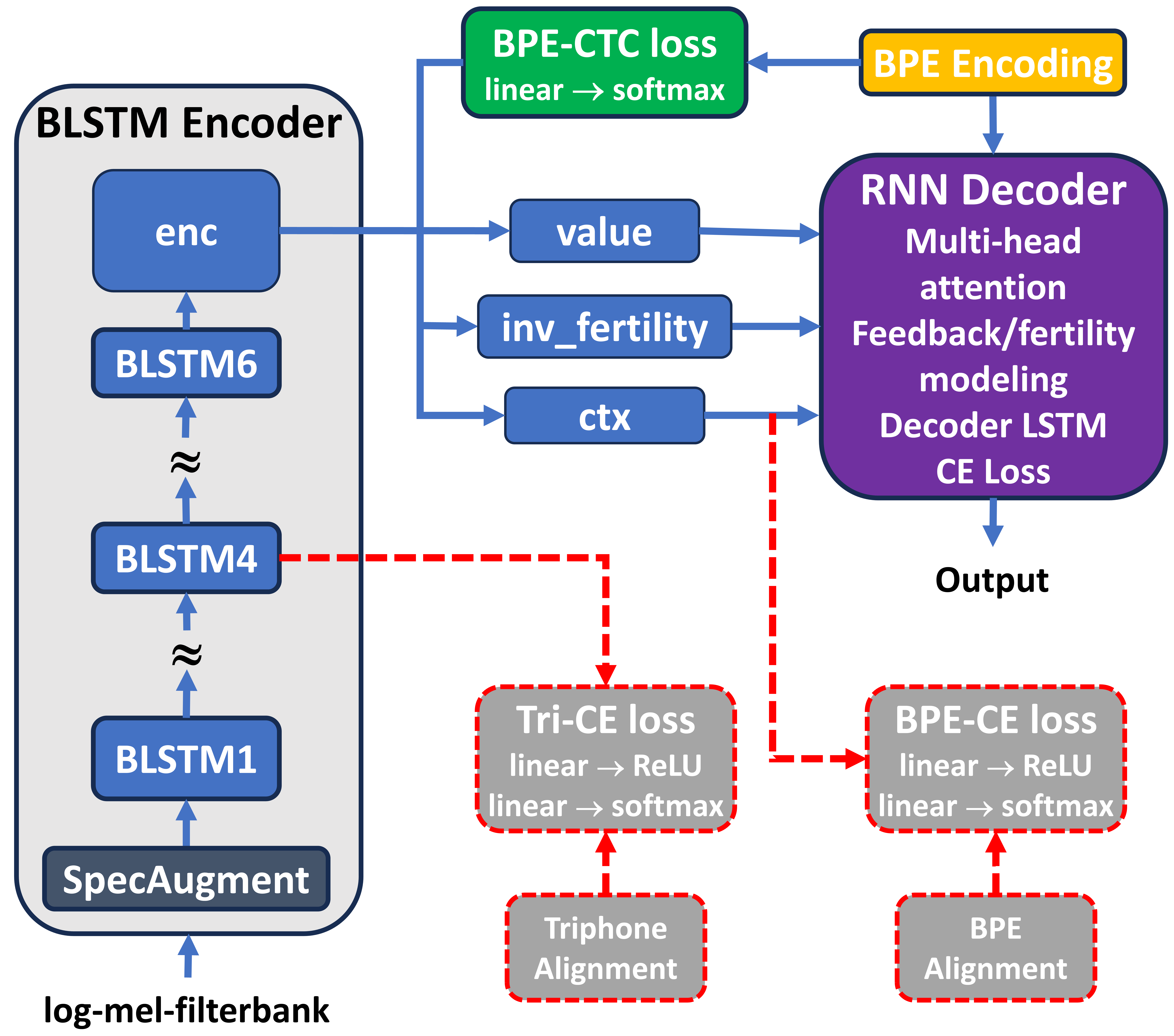}
    \caption{Diagram of the network.}
    \label{fig:diagram}
\end{figure}

\vspace{-.5em}
\label{sec:methodology}\
In the Figure \ref{fig:diagram}, we show a diagram of the proposed AED modeling.
The basic components are the six-layer BLSTM encoder \cite{hochreiter1997long,graves2005bidirectional}, an RNN-based decoder with a LSTM layer inside, and an auxiliary BPE based CTC loss. 
SpecAugment is used to augment the input data \cite{park2019specaugment}.
The encoder output (\texttt{enc}) is projected to encoder value (\texttt{value}), inverse-fertility (\texttt{inv-fertility}), and context (\texttt{ctx}) that are the input to the decoder. 

The dotted red lines/boxes represent the proposed triphone and BPE alignment supervision where a label smoothing of 0.5 is used to construct a weak supervision.
The cross-entropy (CE) based triphone alignment supervision (\texttt{Tri-CE}) is placed at the fourth BLSTM layer (BLSTM4) of the encoder. The rule of thumb is to place the label-smoothed \texttt{Tri-CE} around the the middle layer of the encoder \cite{jiang2023weak,liu2021improving}.
The BPE alignment supervision (\texttt{BPE-CE}) can be placed either at the encoder output (\texttt{enc}) or at the encoder context (\texttt{ctx}). Our previous study \cite{jiang2023weak} shows that it is beneficial to place the BPE-related losses at the high layers of encoder. The \texttt{inv-fertility} layer is not considered here as it is usually optional in the AED networks \cite{zeyer2019comparison}. In initial experiments, we found that the training did not converge when putting both the BPE-CE and BPE-CTC at the encoder output (\texttt{enc}) without additional tuning of the learning rate scheduler.
Given that there are three types of auxiliary tasks (triphone alignment supervision, BPE alignment supervision, and a standard BPE CTC loss), we construct 8 different architectures and losses ($\mathcal{L}$) as defined with the following equations: 
\begin{equation}
\begin{split}
&\mathcal{L}_\text{all} = \mathcal{L}_\text{Primary} + \qquad\qquad\qquad\qquad\quad \mathcal{L}_\text{BPE-CTC-enc} \\
&\mathcal{L}_\text{all} = \mathcal{L}_\text{Primary} + \mathcal{L}_\text{Tri-CE} + \qquad\qquad\quad \mathcal{L}_\text{BPE-CTC-enc} \\
&\mathcal{L}_\text{all} = \mathcal{L}_\text{Primary} + \qquad\qquad \mathcal{L}_\text{BPE-CE-enc} \\
&\mathcal{L}_\text{all} = \mathcal{L}_\text{Primary} + \qquad\qquad \mathcal{L}_\text{BPE-CE-ctx} + \mathcal{L}_\text{BPE-CTC-enc} \\
&\mathcal{L}_\text{all} = \mathcal{L}_\text{Primary} + \qquad\qquad \mathcal{L}_\text{BPE-CE-enc} + \mathcal{L}_\text{BPE-CTC-ctx} \\
&\mathcal{L}_\text{all} = \mathcal{L}_\text{Primary} + \mathcal{L}_\text{Tri-CE} + \mathcal{L}_\text{BPE-CE-enc} \\
&\mathcal{L}_\text{all} = \mathcal{L}_\text{Primary} + \mathcal{L}_\text{Tri-CE} + \mathcal{L}_\text{BPE-CE-ctx} + \mathcal{L}_\text{BPE-CTC-enc} \\
&\mathcal{L}_\text{all} = \mathcal{L}_\text{Primary} + \mathcal{L}_\text{Tri-CE} + \mathcal{L}_\text{BPE-CE-enc} + \mathcal{L}_\text{BPE-CTC-ctx}
\end{split}
\label{eq:loss}
\end{equation}
Where \texttt{enc} and \texttt{ctx} indicate the location for placing the auxiliary losses, and \texttt{Primary} corresponds to the loss of the main task, the CE loss measured at the decoder output. The primary loss is a CE loss with BPEs as targets at the output of the RNN decoder.

In the literature, auxiliary losses are usually scaled to regularize the total loss \cite{lee2021intermediate,liu2021improving}.
As discussed in our previous study \cite{jiang2023weak}, here we apply a label smoothing parameter of 0.5 to create weak supervision, which also leads to better training stability.
In addition to these designs, we also propose alternating the Tri-CE and BPE-CE when both are present. 
The intuition behind the idea is that when there are multiple auxiliary losses running at the same time, there might be conflicting/cancelling effects among them, or they cause learning instability.
The alternating is achieved through the pre-training mechanism in RETURNN \cite{zeyer2018returnn}.
Pre-training is also used in the baseline case to improve convergence stability.

\vspace{-.5em}
\section{Experimental Results}
\vspace{-.5em}
\label{sec:results}

\subsection{Setup}
\vspace{-.5em}
\label{ssec:setup}

The experiments are performed on the TED-LIUM release 2 \cite{rousseau2014enhancing} using RETURNN \cite{zeyer2018returnn}. We follow a recipe from \cite{zeyer2019comparison,jiang2023weak}. In the repository, a full setup is provided. For the TED-LIUM release 2 dataset, we use the same train/dev/test division as in \cite{zeyer2019comparison}. Note that in this study, language models are not applied. 

Triphone alignments are obtained using an existing hybrid BLSTM ASR model that uses 5000 triphones, 80-dimension MFCC as acoustic features, and four 1024-dimension hidden BLSTM layers.
The hybrid model is trained using about 11,000 hours of training data that includes 9893-hour news, 157-hour movies, 961-hour LibriSpeech, and 81-hour WSJ. Traditional Gaussian Mixture Models (GMM) is trained to obtain the triphone CART tree and then the BLSTM model is trained with SpecAugment using RETURNN \cite{zeyer2018returnn}.
The RASR toolkit \cite{rybach2011rasr} is used to perform Viterbi alignments to produces both frame-level triphone and word indices. Here we only use the frame-level triphone indices.

BPE alignments are computed in the following way.
First, each sentence is parsed into BPE tokens.
Then grapheme-to-phoneme (G2P) conversion is applied to derive their pronunciations with three variants allowed.
The \texttt{@@} signs are removed before applying G2P. The G2P model is trained with the Sequitur tool \cite{bisani2008joint}.
After constructing a pronunciation lexicon for BPE tokens, we obtain the BPE alignments the same way as we do for triphone alignments. The aligning process produces both frame-level triphone and BPE indices. Here we only use the frame-level BPE indices.

The structure of the AED model is shown in Figure \ref{fig:diagram}.
The BLSTM layers have a hidden dimension of 1024.
We use dropout and Adam optimizer \cite{kingma2014adam} with a variant of Newbob learning rate scheduling. In the RNN decoder, the CE loss is applied with a label smoothing of 0.1.
A beam size of 12 is used for decoding.
Word Error Rate (WER) is computed on the decoder output using NIST SCLITE Scoring Toolkit \cite{sclite}.

\vspace{-.5em}
\subsection{New Baseline}
\vspace{-.5em}
\label{ssec:newbaseline}
The baseline in \cite{zeyer2018returnn} is quite outdated.
Therefore, we develop a new baseline system and expect the proposed approach can improve on a stronger baseline.

The new baseline is built making the following changes:
\begin{itemize}[leftmargin=*]
  \item Large memory GPU (RTX A6000) is used, and sequence length limitation (maximum BPE length of 75) is removed.
  Previously, 30 hours (about 14\%) of the total audio of 213 hours were removed.
  \item For the training data, we apply a segment boundary non-overlapping extension \footnote{\url{https://github.com/rwth-i6/i6_core/blob/main/datasets/tedlium2.py}} \cite{zeineldeen_chunked_aed}. This will help the AED model for better endpointing.
  \item We increase the number of training sub-epochs from 150 to 200. That equals to 50 full epochs.
  \item We update the implementation of SpecAugment \cite{zeineldeen_chunked_aed}.
  \item We change the input features from 40-dimensional MFCC to 80-dimensional log-mel-filterbank.
  Segment-wise mean and variance normalization is applied.
  \item In the original recipe, the learning rate decay rate of 0.7 is too aggressive.
  Here, we customize learning rate contour and make it decay slowly.
  \item We apply speed perturbation in 20\% of the training time, with randomly chosen factor of 1.1 or 0.9.
  We also add additive noises at raw audio level (20\% of the time and SNR above 0dB) and additive spectral noise to the mel-filterbank (20\% of the time and SNR above 0dB). The noise samples are obtained from Audio Set \cite{gemmeke2017audio}.
  \item Final evaluation model is obtained by averaging the models from the last four sub-epochs (197, 198, 199, and 200) \cite{gao2022revisiting}. 
  \item Other changes include using batch-size of 15000 (frames) instead of 18000 and increasing the sequence ordering buckets.
\end{itemize}

\begin{table}[h!]
\center
\begin{tabular}{|c|c|c|}
\hline
\multirow{2}{*}{Systems} & \multicolumn{2}{c|}{WER [\%]} \\ \cline{2-3} 
      & dev & test \\ \hline \hline
Hybrid ASR with n-gram LM & 10.4 & 9.8 \\ \hline
Baseline in \cite{jiang2023weak} (37.5 epochs) & 13.7 & 10.7 \\ \hline
New-Baseline (50 epochs) & 9.4 & 8.4 \\ \hline
\end{tabular}
\caption{WER improvements from the new baseline.}
\label{tab:newbaseline}
\end{table}

After these steps, we produce a relatively competitive baseline as shown in Table \ref{tab:newbaseline}. We also run the ASR task on the test/dev dataset with the hybrid ASR system (with n-gram LM) used in the triphone/BPE alignment and results are listed in the table. Note that the number of epochs is different in the table. However, this does not change the conclusion that the new baseline is significantly improved as compared to \cite{zeyer2019comparison}.

\subsection{Applying weak triphone and BPE alignment losses}
\vspace{-.5em}
We run a number of experiments according to Equation \ref{eq:loss}. Line 1 in Equation 1 represents the baseline with the original primary loss and BPE CTC loss. Line 2 represents the addition of triphone alignment CE loss (Tri-CE) to the baseline, similar to that in our previous study \cite{jiang2023weak}. Line 3 indicates using BPE alignment CE loss (BPE-CE) to replace the existing BPE CTC loss (BPE-CTC). 
Lines 4 and 5 indicate using both BPE-CE and BPE-CTC at the same time, without Tri-CE.
In these cases, BPE-CE and BPE-CTC losses are either put on \texttt{enc} and \texttt{ctx} representations, respectively, or vice versa.
Lines 6, 7, and 8 are similar to Lines 3, 4, and 5 correspondingly with the addition of the intermediate Tri-CE loss.
For Lines 6-8, we also experiment with alternating the Tri-CE and BPE-CE losses by setting the loss weighting of one to 1.0 and the other to 0.0.
Each pattern is held for five sub-epochs for a total 150 sub-epochs (75\% of training time).
This alternating is achieved through the pre-training mechanism in RETURNN \cite{zeyer2018returnn}.
Note that during decoding, all the loss-related network layers (Tri-CE, BPE-CE, and BPE-CTC) can be removed, and thus all have the same amount of parameters (about 160M). 

Following \cite{zeyer2019comparison}, the learning rates are kept at 0.0008 for 40 sub-epochs and then decay after that (lr1), while for the training of alternating losses, the learning rates are kept at 0.0008 for 150 sub-epochs and then decay after that (lr2). For a comparison purpose, we also apply lr2 to the other experiments.

Experimental results are shown in Table \ref{tab:results}. First, adding triphone alignment CE loss or BPE alignment loss separately improves the ASR results over the baseline. Based on the dev set, it shows that the BPE CE loss produces higher improvement. This can be explained by the fact that the BPE CE loss is consistent with the primary loss as both using BPE tokens. Nevertheless, combination of the triphone CE and BPE CE loss leads to further improvements. When alternating the two CE losses, we see a small degradation on the dev set, while getting the best result on the test set (7.4\%). When applying BPE CE loss, we have choices of putting it on the encoder output (\texttt{enc}), the context output (\texttt{ctx}), or replacing the BPE CTC loss. Based on the dev set, it is favorable to put the BPE CE loss on the encoder output (\texttt{enc}), while moving BPE CTC loss to the context output (\texttt{ctx}). In all cases, removing standard BPE CTC from the pool of auxiliary losses hurts the recognition quality. Overall, the system trained with Tri-CE, BPE-CE on \texttt{enc}, and BPE-CTC on \texttt{ctx} achieves significant performance gain over the baseline (12.8\% and 11.9\% relative improvement for the dev and test sets). The effect of alternating weak triphone/BPE alignment supervision improved test set result slightly further (0.1\% absolute WER gain). The strong learning rates (lr2) work better for the Tri-CE only and BPE-CE only experiments, while the weak learning rates (lr1) work better for the baseline and the Tri-CE and BPE-CE experiments.



\begin{table}[htb]
\centering
\begin{tabular}{|l|c|c|c|c|c|c|c|}
\hline
\multirow{2}{*}{System} & \multirow{2}{*}{\begin{tabular}[c]{@{}c@{}}Tri\\CE\end{tabular}} & \multirow{2}{*}{\begin{tabular}[c]{@{}c@{}}BPE\\CE\end{tabular}} & \multirow{2}{*}{\begin{tabular}[c]{@{}c@{}}BPE\\CTC\end{tabular}} & \multicolumn{2}{c|}{WER (lr1)} & \multicolumn{2}{c|}{WER (lr2)} \\ \cline{5-6} 
       \cline{7-8} & & & & dev & test & dev & test
\\ \hline \hline
Baseline  & - & - & enc & 9.4 & 8.4 & 9.8 & 8.7 \\ \hline
Tri-CE & + &  - & enc & 9.0 & 7.7 & 8.6 & 7.7 \\ \hline
\multirow{3}{*}{BPE-CE} & - & enc & - & 8.7 & 8.0 & 8.5 & 7.7 \\ \cline{2-8} 
& - &  ctx & enc & 8.8 & 7.9 & 8.5 & 7.6 \\ \cline{2-8}
& - & \cellcolor{gray}enc & \cellcolor{gray}ctx & \cellcolor{gray}8.4 & 8.0 & 8.5 & 7.8 \\ \hline
\multirow{3}{*}{\begin{tabular}[l]{@{}c@{}}Tri-CE and\\BPE-CE\end{tabular}} & + & enc & - & 8.2 & 7.6 & 8.6 & 7.7 \\ \cline{2-8}
& + & ctx & enc & 8.1 & 7.5 & 8.3 & 7.4 \\ \cline{2-8}
& + & \cellcolor{gray}enc & \cellcolor{gray}ctx & \cellcolor{gray}8.1 & 7.8 & 8.5 & 7.7 \\ \hline
\multirow{3}{*}{\begin{tabular}[l]{@{}c@{}}Alternating\\Tri-CE and\\BPE-CE\end{tabular}} & + & enc & - & - & - & 8.4 & 7.6 \\ \cline{2-8}
& + & ctx & enc & - & - & 8.3 & 7.5 \\ \cline{2-8}
& + & \cellcolor{gray}enc & \cellcolor{gray}ctx & - & - & \cellcolor{gray}8.2 & 7.4 \\ \hline
\end{tabular}
\caption{WERs [\%] for different auxiliary loss combinations.}
\label{tab:results}
\end{table}


\vspace{-1.5em}

\begin{table}[htb]
\centering
\resizebox{1.01\columnwidth}{!}
{
\begin{tabular}{|@{\hspace{1mm}}l@{}|@{}c@{}|@{}c@{}|@{}c@{}|@{}c@{}|@{}c@{}|c|c|}
\hline
\multirow{2}{*}{\begin{tabular}[l]{@{}c@{}}System\end{tabular}} & \multirow{2}{*}{Enc.} & \multirow{2}{*}{Dec.} & \multirow{1}{*}{\hspace{1mm}Cross\hspace{1mm}} & \multirow{2}{*}{\hspace{1mm}Loss\hspace{1mm}} &
\multirow{2}{*}{\hspace{1mm}Params.\hspace{1mm}} & \multicolumn{2}{c|}{WER [\%]} \\ \cline{7-8} 
       & & & att. & & & dev & test \\ \hline \hline
This work\hspace{1mm} & LSTM & RNN & yes & CE & 160M & 8.2 & 7.4 \\ \hline
 \multirow{2}{*}{ESPnet} & Conf & Transf & yes & CE & 31M & 7.8 & 7.6 \\ \cline{2-8}
                         &\hspace{1mm}EBranch \hspace{1mm}&\hspace{1mm}Transf \hspace{1mm}& yes & CE & 35M & 7.3 & 7.1 \\ \hline
 Citrinet & Conv & - & no & CTC & 142M & 9.8 & 8.9 \\ \hline
\end{tabular}
}
\caption{Overall WER results on TED-LIUM 2 vs. literature.}
\label{tab:comparison}
\end{table}

We also compare the best results here with those from the literature. All results are obtained without language model integration and on TED-LIUM 2 dataset.
Specifically, we compare with the ESPnet results, where the encoder is either a 12-layer conformer or ebranchformer \footnote{\url{https://github.com/espnet/espnet/tree/master/egs2/tedlium2/asr1}} \cite{Gulati2020,kwangyoun2022}.
We also compare with results from Citrinet \cite{majumdar2021citrinet} which is a convolutional CTC model that is trained from scratch. Comparison results are shown in Table \ref{tab:comparison} and it shows that our BLSTM-based AED model produces comparable results with other Conformer-based AED models. However, by no means do we promote BLSTM over Conformer here as usually Conformer produces competitive results, is fast in decoding, and has a smaller memory footprint.
From our experience, Conformer based modeling is more sensitive to the learning rate control and this is more the case when heterogeneous auxiliary losses are involved.
Nevertheless, our future goal is to transfer the improvement we observed here to the state-of-the-art Conformer/Ebranchformer modeling.

\vspace{-.5em}
\section{Conclusion and Discussion}
\vspace{-.5em}
\label{sec:conclusion}
In this work, we proposed to use weak triphone alignment supervision and weak BPE alignment supervision to aid the E2E ASR modeling.
We demonstrated that such alignments extracted by traditional hybrid system complement the standard CTC auxiliary loss and improve cross-attention based encoder-decoder models.
As has been experimentally shown, the proposed weak alignment supervisions resulted in significant gain in speech recognition accuracy, reducing the word error rate by over 10\% relative compared to a standard CTC auxiliary loss based baseline.
Nevertheless, we realize that although the proposed approach works well with the BLSTM encoder, we still need to explore whether the improvement can transfer to transformer-like encoder based AED systems.

\newcommand{\BIBdecl}{\setlength{\itemsep}{0.34 em}}
\bibliographystyle{IEEEtran}
\vspace{-.5em}
\bibliography{strings,refs}

\end{document}